\documentclass[11pt,a4paper]{article}
\usepackage[hyperref]{conll2018}
\usepackage{times}
\usepackage{latexsym}
\usepackage{amssymb}
\usepackage[fleqn]{amsmath}
\usepackage{graphicx}
\usepackage{algorithm}  
\usepackage{algorithmic}
\usepackage{multirow}
\usepackage{array}
\usepackage{pifont}
\graphicspath{{figures/}}
\usepackage{subcaption}
\usepackage[titletoc]{appendix}
\usepackage[utf8]{vietnam}

\usepackage{url}
\usepackage{amsthm}
\usepackage{amsmath}
\usepackage{xspace}

\DeclareMathOperator*{\argmax}{argmax} 

\newcolumntype{L}[1]{>{\raggedright\let\newline\\\arraybackslash\hspace{0pt}}m{#1}}
\newcolumntype{C}[1]{>{\centering\let\newline\\\arraybackslash\hspace{0pt}}m{#1}}
\newcolumntype{R}[1]{>{\raggedleft\let\newline\\\arraybackslash\hspace{0pt}}m{#1}}

\newcommand*\sq{\textsc{Seq2Seq}\xspace}
\newcommand*\msq{\textsc{S2SMix}\xspace}

\newcommand*\reza[1]{{\color{black} #1 }}
\newcommand*\xuanli[1]{{\color{black} #1 }}

\renewcommand{\vec}[1]{\boldsymbol{\mathbf{#1}}}
\def\bx{\vec{x}}
\def\by{\vec{y}}
\def\bh{\vec{h}}
\def\D{\mathcal{D}}
\def\pt{P_{\theta}}
\def\bs{\vec{s}}
\def\bc{\vec{c}}
\def\be{\vec{e}}
\def\ba{\vec{a}}
\def\softmax{\mathrm{softmax}}
\def\bb{\vec{b}}
\def\bv{\vec{v}}

\newcommand\comment[1]{}
\newcommand\secref[1]{Section~\ref{#1}}
\newcommand\figref[1]{Figure~\ref{#1}}
\newcommand\tabref[1]{Table~\ref{#1}}

\def\eg{{\em e.g.,}\xspace}
\def\ie{{\em i.e.,}\xspace}

\newcommand\red[1]{{\color{red} #1}}

\aclfinalcopy 

\title{Sequence to Sequence Mixture Model~~for~~Diverse Machine Translation
}

\author{
  Xuanli He \hspace{2cm} Gholamreza Haffari \hspace{2cm} Mohammad Norouzi\\
  \hspace{1cm} Monash University, Australia \hspace{4cm} Google Brain\\
  {\tt \{xuanli.he1,\,gholamreza.haffari\}@monash.edu} \hspace{.3cm} {\tt mnorouzi@google.com}\\
}

\date{}


\begin{document}
\maketitle

\begin{abstract}
Sequence to sequence (\sq) models often lack diversity in their generated translations. 
This can be attributed to the limitation of \sq models in capturing
lexical and syntactic variations in a parallel corpus resulting from
different styles, genres, topics, or ambiguity of the translation
process.
%
%
In this paper, we develop a novel sequence to  sequence mixture (\msq) model that improves  both translation diversity and quality by adopting a committee of specialized translation models  rather than a single translation model.
Each mixture component selects its own training dataset via  optimization of the marginal log-likelihood, which  leads to a soft clustering of the parallel corpus.
Experiments on four language pairs demonstrate the superiority of our mixture model compared to a \sq baseline with standard or diversity-boosted beam search.
Our mixture model uses negligible additional parameters and incurs no extra computation cost during decoding. 

\end{abstract}

\section{Introduction}

Neural sequence to sequence (\sq) models have been remarkably
effective machine translation (MT) \cite{seq2seq14, dima15}.
They have revolutionized MT by providing a unified end-to-end
framework, as opposed to the traditional approaches requiring several
sub-models and long pipelines.
The neural approach is superior or on-par with
statistical MT in terms of translation quality on
various MT tasks and domains~e.g. \cite{wu2016google,parity2018microsoft}.

A well recognized issue with \sq models is the lack of diversity in the generated translations.
This issue is mostly attributed  to the decoding algorithm \cite{li-diverse-decoder16}, and recently to the model \cite{biao2016,cohnetal2018}. 
The former direction has attempted to design diversity encouraging decoding algorithm, particularly beam search, 
as it generates translations   sharing the majority of their tokens except a few  trailing ones.
The latter direction  has investigated modeling enhancements, particularly the introduction of continuous latent variables, in order to capture lexical and syntactic variations in  training corpora, resulted from the inherent ambiguity of the human translation process.\footnote{For a given source sentence, usually there exist several valid translations.}
%
However, improving the translation  diversity and quality with \sq models is still an open problem, as the results of the aforementioned previous work are not fully satisfactory. 

In this paper, we develop a novel sequence to sequence mixture 
(\msq) model that improves both translation
quality and diversity by adopting a committee of  specialized translation models rather than a single  translation model. 
Each mixture component selects its own training dataset via optimization of  the marginal log-likelihood, which leads to a soft  clustering of the parallel corpus. 
As such, our mixture model introduces a conditioning global discrete latent variable for each sentence, which leads to grouping together and capturing variations in the training corpus.
We design the architecture of \msq such that the mixture components share almost all of their parameters and computation.

We provide experiments on four translation tasks, translating from English to German/French/Vietnamese/Spanish. 
The experiments show that our \msq model consistently outperforms strong baselines, including \sq model with the standard and diversity encouraged beam search, in terms of both translation diversity and quality.
The benefits  of our mixture model comes with negligible additional parameters and no extra computation at inference time, compared to the vanilla \sq model.
%

\section{Attentional Sequence to Sequence}
An attentional sequence to sequence (\sq)
model~\cite{seq2seq14,dima15} aims to directly model the conditional
distribution of an output sequence $\by \equiv (y_1, \ldots, y_T)$
given an input $\bx$, denoted $P(\by \mid \bx)$. This family of
autoregressive probabilistic models decomposes the output distribution
in terms of a product of distributions over individual tokens, often
ordered from left to right as,
\begin{equation}
  \pt(\by \mid \bx) ~=~ \prod\nolimits_{t=1}^{\lvert \by \rvert} \pt(y_t \mid \by_{<t}, \bx)~,
\label{equ:1}
\end{equation}
where $\by_{<t} \equiv (y_1, \ldots, y_{t-1})$ denotes a prefix of the
sequence $\by$, and $\theta$ denotes the tunable parameters of the
model.

Given a training dataset of input-output pairs, denoted by $\D \equiv
\{(\bx, \by^*)_d\}_{d=1}^D$, the {\em conditional log-likelihood} objective,
predominantly used to train \sq models, is expressed as,
\begin{equation}
\ell_{\text{CLL}}(\theta) = \!\!\! \sum_{(\bx, \by^*) \in \D}\sum_{t=1}^{\lvert \by^* \rvert} \log \pt(y^*_t \mid \by^*_{<t}, \bx)~.
\label{equ:3}
\end{equation}

\comment{
\begin{figure*}[ht]
\includegraphics[scale=0.35]{seq2seq.png}
\centering
\caption{\sq model with attention schema}
\label{fig:1}
\end{figure*}
}

A standard implementation of the \sq model is composed of an encoder
followed by a decoder. The encoder transforms a sequence of source
tokens denoted $(x_1,\ldots,x_N)$, into a sequence of hidden states
denoted $(\bh_1,\ldots,\bh_N)$ via a recurrent neural network
(RNN). Attention provides an effective mechanism to represent a soft
alignment between the tokens of the input and output
sequences~\cite{dima15}, and more recently to model the dependency
among the output variables~\cite{transformer17}.

In our model, we adopt a bidirectional RNN with LSTM units
\cite{hochreiter1997long}. Each hidden state $\bh_n$ is the
concatenation of the states produced by the forward and backward RNNs,
$\bh_n=[\bh_{\rightarrow n}, \bh_{n\leftarrow}]$. Then, we use a two-layer
RNN decoder to iteratively emit individual distributions over
target tokens $(y_1,...,y_T)$.
At time step $t$, we compute the hidden representations of an output
prefix $\by_{\le t}$ denoted $\bs_t^{1}$ and $\bs_t^{2}$ based on an
embedding of $y_t$ denoted $\mathrm{M}[y_t]$, previous representations
$\bs_{t-1}^{1}$, $\bs_{t-1}^{2}$, and a context vector $\bc_t$ as,
\begin{eqnarray}
\label{eq:lstm1}
\hspace{-.6cm}\bs_t^{1}                &\!\!\!=\!\!\!& \mathrm{LSTM}(\bs_{t-1}^{1}, \mathrm{M}[y_t];\bc_t)~,\\
\label{eq:lstm2}
\hspace{-.6cm}\bs_t^2                 &\!\!\!=\!\!\!& \mathrm{LSTM}(\bs_{t-1}^2, \bs_{t}^1;\bc_t)~,\\
\label{eq:softmax}
\hspace{-.6cm}\pt(y_{t+1}\!\mid\!\by_{\le t},\bx) &\!\!\!=\!\!\!& \softmax(W\bs_t^2 + W'\bc_t)~,
\label{equ:2}
\end{eqnarray}
where $\mathrm{M}$ is the embedding  table, and $W$ and $W'$ are learnable parameters. 
The  context vector 
$\bc_t$ is  computed based on the input and attention,
\begin{eqnarray}
\hspace{-.6cm}e_{t,n}&\!\!=\!\!&\bv^{\top}\mathrm{tanh}(W_h\bh_n+W_s\bs_{t-1}^1+\bb_{a})~,\\
\hspace{-.6cm}\ba_t &\!\!=\!\!& \softmax(\be_t)~,\\
\hspace{-.6cm}\bc_t&\!\!=\!\!&\sum_na_{t,n}\,\bh_n~,
\end{eqnarray}
where $W_h$, $W_s$, $\bb_{a}$, and $\bv$ are learnable parameters, and
$\ba_t$ is the attention distribution over the input tokens at time
step $t$. The decoder utilizes the attention information to decide
which input tokens should influence the next output token $y_{t+1}$.

\section{Sequence to Sequence Mixture Model}
\label{sec:mixture}
We develop a novel {\em sequence to sequence mixture} (\msq) model
that improves both translation quality and diversity by adopting a
committee of specialized translation models rather than a single
translation model. Each mixture component selects its own training
dataset via optimization of the marginal log-likelihood, which leads
to a soft clustering of the parallel corpus. We design the
architecture of \msq such that the mixture components share almost all
of their parameters except a few conditioning parameters. This enables
a direct comparison against a \sq baseline with the same number of
parameters.

Improving translation diversity within \sq models has received
considerable recent attention
(\eg~\citet{vijayakumar2016diverse,li-diverse-decoder16}).  Given a
source sentence, human translators are able to produce a set of
diverse and reasonable translations. However, although beam search
for \sq models is able to generate various candidates, the final
candidates often share majority of their tokens, except a few trailing
ones. The lack of diversity within beam search raises an issue for
possible re-ranking systems and for scenarios where one is willing to
show multiple translation candidates to the user. Prior work attempts
to improve translation diversity by incorporating a diversity penalty
during beam
search~\cite{vijayakumar2016diverse,li-diverse-decoder16}. By
contrast, our \msq model naturally incorporates diversity both during
training and inference.

The key difference between the \sq and \msq models lies in the
formulation of the conditional probability of an output sequence $\by$
given an input $\bx$. The \msq model represents $\pt(\by \!\mid\! \bx)$ by
marginalizing out a discrete latent variable $z \in \{1, \ldots, K\}$,
which indicates the selection of the mixture component, \ie~
\begin{equation}
\pt(\by \!\mid\! \bx) = \sum_{z=1}^{K} \pt(\by \!\mid\! \bx,z)\,P(z\!\mid\!\bx)~,
\label{eq:mixp}
\end{equation}
where $K$ is the number of mixture components. For simplicity and to
promote diversity, we assume that the mixing coefficients follow a uniform
distribution such that for all $z \in \{1, \ldots, K\}$,
\begin{equation}
P(z \mid \bx)~=~{1}/{K}~.
\label{eq:unimix}
\end{equation}

For the family of \msq models with uniform mixing
coefficients \eqref{eq:unimix}, the conditional log-likelihood
objective~\eqref{equ:3} can be re-expressed as:
\begin{equation}
\begin{aligned}
\ell&_{\text{CLL}}(\theta) ~=~ \mathrm{constant} ~+\\
&\sum_{(\bx, \by^*) \in \D}\log\sum_{z=1}^K\underbrace{\exp\sum_{t=1}^{\lvert \by^* \rvert} \log \pt(y^*_t \mid \by^*_{<t}, \bx,z)}_{\pt(\by \mid \bx,z)}~,
\end{aligned}
\label{eq:s2smixloss}
\end{equation}
where $\log(1/K)$ terms were excluded because they offset the
objective by a constant value. Such a constant has no impact on
learning the parameters $\theta$. One can easily implement the
objective in \eqref{eq:s2smixloss} using automatic differentiation
software such as tensorflow~\cite{tf16}, by adopting a
$\mathrm{LogSumExp}$ operator to aggregate the loss of the individual
mixture components. When the number of components $K$ is large,
computing the terms $\pt(y^*_t \mid \by^*_{<t}, \bx,z)$ for all values
of $z\in\{1,\ldots,K\}$ can require a lot of GPU memory. To mitigate
this issue, we will propose a memory efficient formulation
in \secref{sec:mem-efficient} inspired by the EM algorithm.

\subsection{\msq Architecture}

\begin{figure}[t]
\includegraphics[scale=0.3]{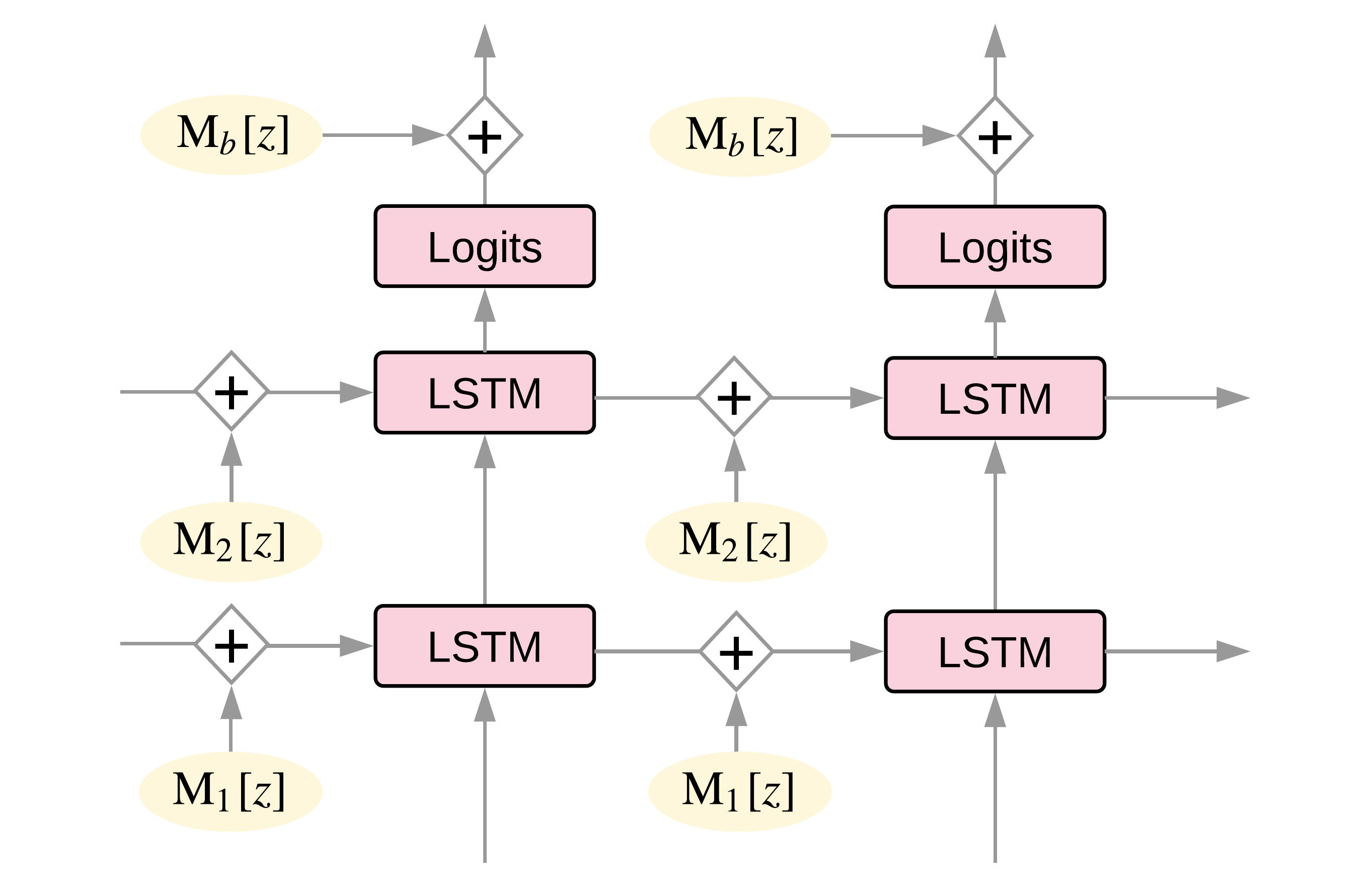}
\caption{An illustration of a two-layer LSTM decoder with different ways of injecting the conditioning signal.}
\label{fig:2}
\end{figure}

We design the architecture of the \msq model such that individual
mixture components can share as many parameters and as much
computation as possible. Accordingly, all of the mixture components
share the same encoder, which requires processing the input sentence
only once. We consider different ways of injecting the conditioning
signal into the decoder. As depicted in \figref{fig:2}, we consider
different ways of injecting the conditioning on $z$ into our two-layer
decoder. These different variants require additional lookup tables
denoted $\mathrm{M}_1, \mathrm{M}_2$, or $\mathrm{M}_b$.

When we incorporate the conditioning on $z$ into the LSTM layers, each
lookup table (\eg~$\mathrm{M}_1$ and $\mathrm{M}_2$) has $K$ rows and
$D_{\text{LSTM}}$ columns, where $D_{\text{LSTM}}$ denotes the number
of dimensions of the LSTM states ($512$ in our case). We combine the
state of the LSTM with the conditioning signal via simple
addition. Then the LSTM update equations take the form,
\begin{equation}
\bs_t^{i}  ~=~ \mathrm{LSTM}(\bs_{t-1}^{i} + \mathrm{M}_i[z],~\mathrm{input})~,
\label{eq:lstm2}
\end{equation}
for $i \in \{1, 2\}$. We refer to the addition of the conditioning
signal to the bottom and top LSTM layers of the decoder as {\em bt}
and {\em tp} respectively. Note that in the bt configuration, the
attention mask depends on the indicator variable $z$, whereas in the
tp configuration that attention mask is shared across different
mixture components.

We also consider incorporating the conditioning signal into the
softmax layer to bias the selection of individual words in each
mixture component. Accordingly, the embedding table $\mathrm{M}_b$ has
$K$ rows and $D_{\text{vocab}}$ entries, and the logits
from \eqref{eq:softmax} are added to the corresponding row of
$\mathrm{M}_b$ as,
\begin{equation}
\pt(y_{t+1}\!\mid\!\by_{\le t},\bx,z) ~=~ \softmax(\mathrm{logits} + \mathrm{M}_b[z])~.
\end{equation}
We refer to this configuration as {\em sf} and to the configuration
that includes all of the conditioning signals as {\em all}.

\subsection{Separate Beam Search per Component}\label{sec:sbs}

At the inference stage, we conduct a separate beam search per mixture
component. Performing beam search independently for each component
encourages diversity among the translation candidates as different
mixture components often prefer certain phrases and linguistic
structures over each other. Let $\hat{\by}_z$ denote the result of the
beam search for a mixture component $z$. The final output of our
model, denoted $\hat{\by}$ is computed by selecting the translation
candidate with the highest probability under the corresponding
mixture component, \ie~
\begin{equation}
\hat{\by}~=~\argmax_{1 \le z \le K}~\log \pt(\hat{\by}_z\mid \bx,z)~.
\label{eq:infm}
\end{equation}

In order to accurately estimate the conditional probability of each
translation candidate based on \eqref{eq:mixp}, one needs to evaluate
each candidate using all of the mixture components. However, this
process considerably increases the inference time and
latency. \reza{Instead, we approximate the probability of each candidate by
only considering the mixture component based on which the
candidate translation has been decoded, as outlined in \eqref{eq:infm}. This approximation also encourages the diversity
as we emphasized in this work.}


Note that we have $K$ mixture components and a beam search of $b$ per
component. Overall, this requires processing $K\times b$
candidates. Accordingly, we compare our model with a \sq model using
the same beam size of $K\times b$.

\subsection{Memory Efficient Formulation}
\label{sec:mem-efficient}

In this paper, we adopt a relatively small number of mixture
components (up to $16$), but to encompass various clusters of
linguistic content and style, one may benefit from a large number of
components. Based on our experiments, the memory footprint of a \msq
with $K$ components increases by about $K$ folds, partly because the
softmax layers take a big fraction of the memory. To reduce the memory
requirement for training our model, inspired by prior work on EM
algorithm~\cite{em98}, we re-express the gradient of the conditional
log-likelihood objective in~\eqref{eq:s2smixloss} {\em exactly} as,
\begin{equation}
\begin{aligned}
&\frac{d}{d\theta}\ell_{\text{CLL}}(\theta) ~=~\\
&\sum_{(\bx, \by^*) \in \D}\sum_{z=1}^K P(z\mid \bx,\by^*) \frac{d}{d\theta} \log \pt(\by^* \mid \bx,z)~,
\end{aligned}
\label{eq:grad-cllm}
\end{equation}
where with uniform mixing coefficients, the posterior distribution
$P(z\mid \bx,\by^*)$ takes the form,
\begin{equation}
P(z\mid \bx,\by) ~=~\frac{\exp\ell_z(\by\!\mid\!\bx)}{\sum_k \exp\ell_k(\by\!\mid\!\bx)}~,
\label{eq:s2s-pos}
\end{equation}
where $\ell_z(\by\!\mid\!\bx) = \log \pt(\by \!\mid\! \bx,z)$.

Based on this formulation, one can compute the posterior distribution
in a few forward passes, which require much less memory. Then, one can
draw one or a few Monte Carlo (MC) samples from the posterior to
obtain an unbiased estimate of the gradient in \eqref{eq:grad-cllm}.
As shown in algorithm \ref{alg:1}, the training procedure is divided
into two parts. For each minibatch we compute the component-specific
log-loss for different mixture components in the first stage. Then, we
exponentiate and normalize the losses as in \eqref{eq:s2s-pos} to
obtain the posterior distribution. Finally, we draw one sample from
the posterior distribution per input-output example, and optimize the
parameters according to the loss of such a component. These two stages
are alternated until the model converges. We note that this algorithm
follows an {\em unbiased} stochastic gradient of the marginal log
likelihood.

\begin{algorithm} 
\caption{Memory efficient \msq} 
\label{alg:1} 
\begin{algorithmic} 
\STATE Initialize a computational graph: cg
  \STATE {Initialize a optimizer: opt}
    \REPEAT
         \STATE{draw a random minibatch of the data}
         \STATE{empty list $\Gamma$}
         \FOR{$z=1$ \TO $K$}
         \STATE {$\ell_z$ := cg.forward(minibatch, $z$)}
         \STATE{${\Gamma}$ := add $\exp(\ell_z)$ to $\Gamma$ }
           \ENDFOR
           \STATE{$\Gamma$ := normalize(${\Gamma}$)}
         \STATE {$\tilde{z}$ := sample($\Gamma$)}
         \STATE {$\ell$ := cg.forward(minibatch, $\tilde{z}$)}
         \STATE{opt.gradient\_descent($\ell$)}
     \UNTIL{converge}
\end{algorithmic}
\end{algorithm}

\section{Experiments}
\paragraph{Dataset.}
To assess the effectiveness of the \msq model, we conduct a set of
translation experiments on TEDtalks on four language pairs:
English$\to$French (en-fr), English$\to$German (en-de),
English$\to$Vietnamese (en-vi), and English$\to$Spanish (en-es).
%

We use IWSLT14 dataset\footnote{https://sites.google.com/site/iwsltevaluation2014/home}
for en-es, IWSLT15 dataset for en-vi, and IWSLT16
dataset\footnote{https://sites.google.com/site/iwsltevaluation2016} for
en-fr and en-de.
We pre-process the corpora by Moses
tokenizer\footnote{https://github.com/moses-smt/mosesdecoder}, and
preserve the true case of the text.
For en-vi, we use the pre-processed corpus distributed
by \citet{luong2015stanford}\footnote{https://nlp.stanford.edu/projects/nmt}.
For training and dev sets, we discard all of the sentence pairs where
the length of either side exceeds $50$ tokens.
The number of sentence pairs of different language pairs after
preprocessing are shown in
\tabref{tab:1}.
We apply byte pair encoding (BPE)~\cite{sennrich2016neural} to handle
rare words on en-fr, en-de and en-es, and share the BPE vocabularies
between the encoder and decoder for each language pair.
%

\begin{table}[t]
\begin{center}
\begin{tabular}{ccccc}
 \hline    
    Data& en-fr  &en-de& en-vi & en-es\\
    \hline
        Train & 208,719 & 189,600 &133,317&173,601 \\
         Dev&   5,685 &6,775&1,553&5,401\\
        Test & 2,762 & 2,762 &1,268&2,504\\
         \hline
\end{tabular}
\end{center}
\caption{Statistics of all language pairs for IWSLT data after preprocessing}
\label{tab:1}
\end{table}


\paragraph{Implementation details.} 
All of the models use a one-layer bidirectional LSTM encoder and a
two-layer LSTM decoder. Each LSTM layer in the encoder and decoder has
a $512$ dimensional hidden state. Each input word embeddings is $512$
dimensional as well. We adopt the Adam
optimizer~\cite{DBLP:journals/corr/KingmaB14}.  We adopt dropout with
a $0.2$ dropout rate. The minibatch size is $64$ sentence pairs.  We
train each model $15$ epochs, and select the best model in terms of
the perplexity on the dev set.

\paragraph{Diversity metrics.} 
Having more diversity in the candidate translations is one of the
major advantages of the \msq model.
To quantify diversity within a set $\{\hat{\by}_m\}_{m=1}^M$ of
translation candidates, we propose to evaluate average pairwise BLEU
between pairs of sentences according to
{
\begin{equation}
\textrm{div\_bleu}\equiv 100 - \frac{\displaystyle\sum_{i=1}^{M} \sum_{j={i+1}}^{M} \textrm{BLEU}(\hat{\by}_i,\hat{\by}_j)}{M(M-1)/2} 
\end{equation}
}

As an alternative metric of diversity within a set
$\{\hat{\by}_m\}_{m=1}^M$ of translations, we propose a metric based
on the fraction of the n-grams that are not shared among the
translations, \ie~
\begin{equation}
\textrm{div\_ngram} \equiv 1-
\frac{\left\lvert \displaystyle\cap^{{M}}_{m=1}\textrm{ngrams}(\hat{\by}_m) \right\rvert}
{~\left\lvert \displaystyle\cup^{{M}}_{m=1}\textrm{ngrams}(\hat{\by}_m) \right\rvert~}
\end{equation}
where $\textrm{ngram}(\by)$ returns the set of unique n-grams in a
sequence $\by$. We report average div\_bleu and average div\_ngram
across the test set for the translation candidates found by beam search.
We measure and report bigram diversity in the paper
and report unigram diversity in the supplementary material.

\subsection{Experimental results}

\begin{figure}[t]
\includegraphics[width=7cm]{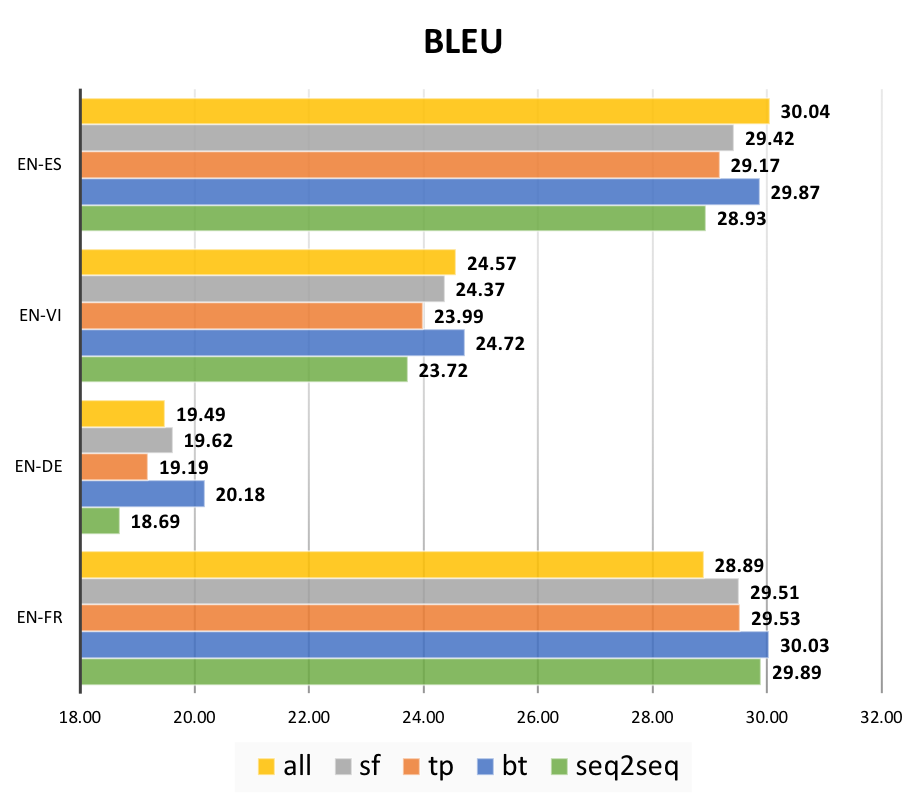}
\caption{\xuanli{BLEU scores of the different variants of \msq model and \sq model.}}
\label{fig:variants}
\end{figure}

\paragraph{\msq configuration.} 

We start by investigating which of the ways of injecting the
conditioning signal into the \msq model is most effective.
As seen in \secref{sec:mixture}, the mixture components can be built
by adding component-specific vectors to the logits (sf), the top LSTM
layer (tp) or the bottom LSTM layer (bt) in the decoder, or all of
them (all).
\figref{fig:variants} shows the BLEU score of these variants on
the translation tasks across four different language pairs.
%
We observe that adding a component-specific vector to the recurrent
cells in the bottom layer of the decoder is the most effective, and
results in BLEU scores superior or on-par with the other variants
across the four language pairs.

Therefore, we use this model variant in all experiments for the rest of the paper. 

Furthermore, Table \ref{tab:6} shows the number of parameters in each
of the variants as well as the base \sq model.
We confirm that the mixture model variants introduce negligible number
of new parameters compared to the base \sq model. 
Specifically, only up to $0.002\%$ increase in the parameter size are
introduced, across all of the language pairs and mixture model variants.

\begin{table}[t]
\begin{center}
\setlength\tabcolsep{4pt}
\begin{tabular}{lcccc}
 \hline    
    &en-fr&en-de &  en-vi&  en-es\\
         \hline
         \sq & 173.22 & 173.78 &112.76&173.21  \\
          \msq-4& & & & \\
          \ \ \   bt & 173.23 & 173.79 &112.77&173.22 \\
           \ \ \  tp  & 173.23 & 173.79 &112.77&173.22\\
          \ \ \  sf & 173.70 &  174.27&112.88&173.70\\
          \ \ \  all & 173.72 & 174.29 &112.90&173.72\\
        \hline
\end{tabular}
\end{center}
\caption{Size of the parameters (MB) for the base \sq\ model and the variants of \msq with four mixtures.}
\label{tab:6}
\end{table}

\paragraph{\msq vs. \sq.}    
We compare our mixture model against a vanilla \sq model both in terms
of translation quality and diversity.
To be fair, we compare models with the same number of beams during
inference, \eg~we compare vanilla \sq using a beam size of $8$
with \msq-4 with $4$ component and a beam size of $2$ per component.
%

\begin{table}[t]
\begin{center}
\setlength\tabcolsep{4pt}
\begin{tabular}{c c | c c c c }
  & beam &en-fr& en-de&en-vi & en-es\\
   \hline  \hline 
   \multicolumn{1}{ c  }{\multirow{ 3}{*}{\sq}}&{4}&30.26 &19.52&24.82&29.40\\
   \multicolumn{1}{ c  }{}&{8}&30.18 &19.77&23.55&29.76\\
   \multicolumn{1}{ c  }{}&{16}&27.63 &19.13&19.05&28.19\\
   \hline
  \multicolumn{1}{ c  }{\multirow{ 3}{*}{\msq-4}}&{1}&30.61 &20.18&25.16&31.17\\
  \multicolumn{1}{ c  }{} &{2}&31.22 &20.71&25.28&31.47\\
   \multicolumn{1}{ c  }{}&{4}&31.97 &21.08&25.36&31.21\\
 
\end{tabular}
\end{center}
\caption{BLEU scores of different systems over different search space.}
\label{tab:main}
\end{table}

\xuanli{As an effective regularization strategy, we adopt label smoothing  to strengthen generalisation 
performance \cite{szegedy2016rethinking,pereyra2017regularizing,sergeynaacl18}. Unlike conventional cross-entropy loss, where the probability mass  for the ground truth word $y$ is set to 1 and $q(y')=0$ for $y'\neq y$, we
smooth this distribution as:
\begin{eqnarray}
\hspace{-.6cm}q(y)&\!\!=\!\!&1-\epsilon,\\
\hspace{-.6cm}q(y') &\!\!=\!\!&\frac{\epsilon}{V-1}
\end{eqnarray}
where $\epsilon$ is a smoothing parameter, and $V$ is the vocabulary size. 
In our experiments, $\epsilon$ is set to 0.1.}

Table \ref{tab:main} shows the results across four language
pairs. Each row in the top part should be compared with the
corresponding row in the bottom part for a fair comparison in terms of
the effective beam size.
Firstly, we observe that  increasing the beam size deteriorates the BLEU score for the \sq\ model.
Similar observations have been made in the previous work \cite{tu2017neural}.
This behavior is in contrast to our \msq\ model where increasing the  beam size improves the BLEU score, except
en-es, which demonstrates a decreasing trend when beam size increases from 2 to 4. 
Secondly, our \msq\ models outperform their \sq\ counterparts in all settings with the same number of bins. 
%

Figure \ref{fig:div} shows the diversity comparison between the \msq\ model and the vanilla \sq\ model where the number of decoding beams is 8. 
The diversity metrics are bigram and BLEU diversity as defined earlier in the section. 
Our \msq\ models significantly dominate the \sq\ model across language pairs in terms of the diversity metrics, while keeping the translation quality high (c.f. the BLEU scores in Table \ref{tab:main}). 
%

We further compare against the \sq \ model endowed with the beam-diverse decoder \cite{li-diverse-decoder16}.  
This decoder penalizes sibling hypotheses generated from the same parent in the search tree, according to their ranks in each decoding step.  
Hence, it tends to rank high those hypotheses from different parents, hence encouraging diversity in the beam. 

Table \ref{tab:div} presents the BLEU scores as well as the diversity measures. 
As seen, the mixture model significantly outperforms the \sq\ endowed with the  beam-diverse decoder, in terms of the diversity in the generated translations. Furthermore, the mixture model achieves up to 1.7 BLEU score improvements across three language pairs.

\begin{table}[t]
\begin{center}
\begin{tabular}{l||c|c|c|c}
                 &en-fr&en-de& en-vi& en-es\\
         \hline \hline
         BLEU & & & & \\
          \ \ \ \sq-d & 29.85& 19.18& 24.62&29.72  \\
          \ \ \ \msq-4 & 30.61 &20.18&25.16&31.17 \\
                    \hline
          DIV\_BLEU & & & &  \\
         \ \ \ \sq-d & 20.43& 22.66& 14.51&18.83  \\
          \ \ \ \msq-4 & 34.85& 47.85& 37.40&38.31 \\
\end{tabular}
\end{center}
\caption{\msq\ with 4 components vs \sq\ endowed with the beam-diverse decoder \cite{li-diverse-decoder16} with the beam size of 4.}
\label{tab:div}
\end{table}

\begin{table*}[t]
\begin{center}
\begin{tabular}{l||cc|cc|cc|cc}
                 &\multicolumn{2}{c}{en-fr}&\multicolumn{2}{c}{en-de} &  \multicolumn{2}{c}{en-vi}&  \multicolumn{2}{c}{en-es}\\
                 & BLEU & time & BLEU & time & BLEU & time & BLEU & time \\
         \hline \hline
         \msq-4 
                &30.61 &1.25 &20.18 & 1.33& 25.16&1.14 &31.17 &1.30 \\
          MC sampling: & & & & & & & & \\
         \ \ \ \ \msq-4 & 30.43& 1.67& 19.74&1.67 & 24.93&1.58 &31.27 & 1.67\\
         \ \ \ \ \msq-8 &30.66 & 2.08&20.41 & 2.05& 24.86& 2.00& 31.44& 2.06\\
         \ \ \ \ \msq-16 &30.74 &3.10 &20.43 &2.88 & 24.90&2.83 & 30.82&3.02 \\
\end{tabular}
\end{center}
\caption{BLEU scores using greedy decoding and training time based
  on the original log-likelihood objective and online EM coupled with
  gradient estimation based on a single MC sample. The training time is
  reported by taking the average running time of one minibatch update
  across a full epoch.}
\label{tab:em}
\end{table*}

\paragraph{Large mixture models.}  
Memory limitations of the GPU may make it difficult
to increase the number of mixture components beyond a certain amount.
One approach is to decrease the number of sentence pairs in a minibatch,
however, this results in a substantial increase in the training time.
Another approach is to resort to MC gradient estimation as discussed
in \secref{sec:mem-efficient}.
%

%
The top-part of Table \ref{tab:em} compares the models trained by
online EM vs the original log-likelihood objective, in terms of the
BLEU score and the training time. As seen, the BLEU score of the
EM-trained models are on-par with those trained on the log-likelihood
objective. However, online EM leads to up to 35\% increase in the
training time for \msq-4\ across four different language pairs, as we
first need to do a forward pass on the minibatch in order to form the
lower bound on the log-likelihood training objective.

The bottom-part of Table \ref{tab:em} shows the effect of online EM coupled with sampling only one mixture component to form a stochastic approximation to the log-likelihood lower bound.  
For each minibatch, we first run a forward pass to compute the probability of each mixture component given each sentence pair in the minibatch. We then sample a mixture component for each sentence-pair to form the approximation of the log-likelihood lower bound for the minibatch, which is then optimized using back-propagation. 
 \xuanli{As we increase the number of mixture components from 4 to 8, we see about 0.7 BLEU score increase for en-de; while there is no significant change in the BLEU score for en-fr, en-vi and en-es. }

 Increasing the number of mixture components further to 16 does not produce gains on these datasets. 
 Time-wise, training with online EM coupled with 1-candidate sampling should be significantly faster that the vanilla online 
 EM and  the original likelihood objective in principle, as we  need to perform the backpropagation only for the 
 selected mixture component (as opposed to all mixture components).
 Nonetheless, the additional computation due to increasing the number of mixtures from 4 to 8 is about 26\%, which increases to about 
 55\% when going from 8 to 16 mixture components.
 
\begin{table*}[h]
\begin{center}
\begin{tabular}{lll|}
\hline
\hline
Source & And this information is stored for at least six months in Europe , up to two years .\\
Reference &Và \underline{những} thông tin này được lưu trữ trong ít nhất \underline{sáu} tháng ở châu Âu , \underline{cho tới tận hai} năm .\\
\sq&\\
&Và thông tin này được lưu trữ trong ít sáu tháng ở Châu Âu , hai năm tới .\\
&Và thông tin này được lưu trữ trong ít sáu tháng ở Châu Âu , trong hai năm tới .\\
&Và thông tin này được lưu trữ trong ít sáu tháng ở Châu Âu , hai năm tới\\
&Và thông tin này được lưu trữ trong ít sáu tháng ở Châu Âu , trong hai năm tới\\

\msq&\\
&Và thông tin này được lưu trữ trong ít \red{nhất} \underline{6} tháng ở châu Âu , \red{\underline{đến 2}} năm .\\
&Và thông tin này được lưu trữ trong ít \red{nhất} \underline{6} tháng ở châu Âu , \red{\underline{lên tới hai}} năm .\\
&Và thông tin này được lưu trữ trong ít \red{nhất} \underline{6} tháng ở châu Âu , \red{\underline{trong vòng hai}} năm .\\
&Và thông tin này được lưu trữ trong ít \red{nhất} \underline{6} tháng ở châu Âu , \red{\underline{lên tới hai}} năm\\
\hline\hline
\end{tabular}
\end{center}
\caption{ \underline{Words} indicate diversity compared with the references, while \red{red} words denote translation improvement.}
\label{tab:qual}
\end{table*}

\begin{figure}[h]
    \centering
    \begin{subfigure}[b]{0.40\textwidth}
        \includegraphics[width=\textwidth]{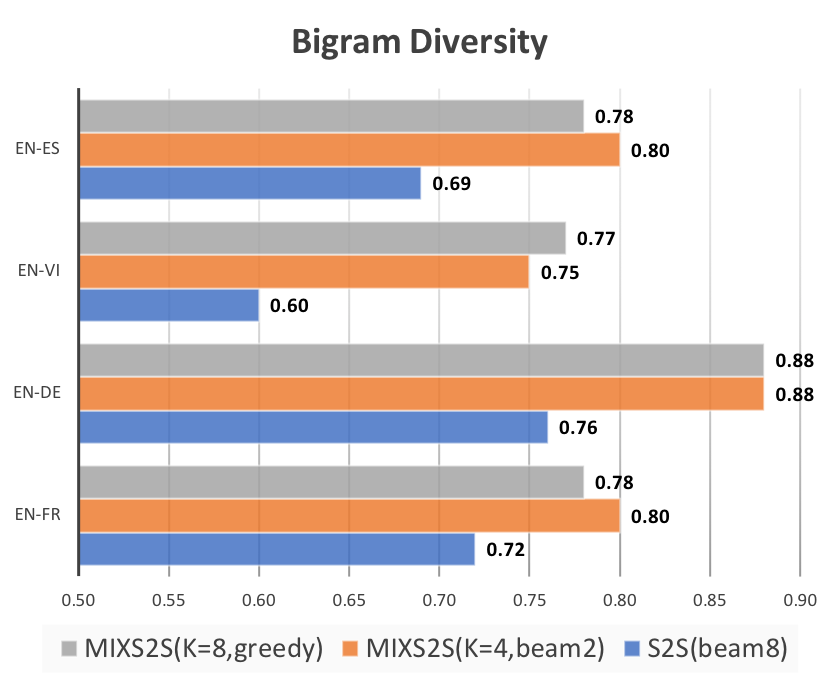}
    \end{subfigure}
    ~ 
    \begin{subfigure}[b]{0.40\textwidth}
        \includegraphics[width=\textwidth]{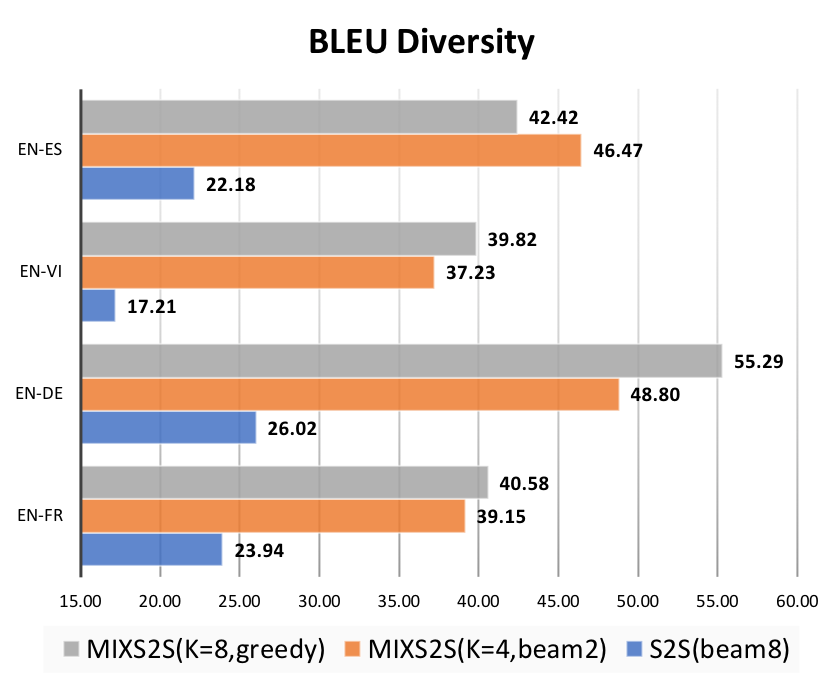}
    \end{subfigure}
    \caption{Diversity bigram (top) and BLEU (bottom) for the  \sq\ model vs \msq models, with the  number of decoding beams  set to 8.}
    \label{fig:div}
\end{figure}

\subsection{Qualitative Analysis}
Finally, we would like to demonstrate that our \msq does indeed encourage diversity and improve the translation quality. As shown in \tabref{tab:qual}, compared with \sq, which mistranslates the second
clause, our \msq is not only capable of generating a group of correct translation, but also emitting synonyms for different mixture components. We provide more examples in the supplementary material.

\section{Related Work}
Obviously, different domains aim at different readers, thus they exhibit distinctive genres compared to other 
domains. A well-tuned MT system cannot directly apply to new domains; otherwise, translation quality will 
degrade. Based on this factor, out-domain adaptation has been widely studied for MT, ranging from data selection 
\cite{li2010adaptive,wang2017instance}, tuning \cite{luong2015stanford,farajian2017multi} to domain tags 
\cite{DBLP:journals/corr/ChuDK17}. Similarly, in-domain adaptation is also a compelling direction. Normally, to
train an universal MT system, the training data consist of gigantic corpora covering numerous and various
domains.This training data is naturally so diverse that \citet{mima1997improving} incorporated 
extra-linguistic information to enhance translation quality. \citet{michel18acl} argue even 
without explicit signals (gender, politeness etc.), they can handle domain-specific information via annotation
of speakers, and easily gain quality improvement from a larger number of domains. Our approach is considerably
different from the previous work. We remove any extra annotation, and treat domain-related information as latent
variables, which are learned from corpus. 

Prior to our work, diverse generation has been studied in image captioning, as some of the training set are 
comprised of images paired with multiple reference captions. Some work puts their efforts on decoding stages, and 
form a group of beam search to encourage diversity \cite{vijayakumar2016diverse}, while others pay more attention
to adversarial training \cite{shetty2017speaking,li2018generating}. Within translation, our method is similar to
\citet{philip18acl}, where they propose a MT system armed with variational inference to account 
for translation variations. Like us, their diversified generation is driven by latent variables. Albeit the 
simplicity of our model, it is effective and able to accommodate variation or diversity. Meanwhile, we propose 
several diversity metrics to perform quantitative analysis.

\xuanli{Finally, \citet{yang2018breaking} proposes a mixture of softmaxes  to enhance the expressiveness of language model, which demonstrate the effectiveness of our \msq model under the matrix factorization framework.}

\section{Conclusions and Future Work}
In this paper, we propose a sequence to sequence mixture (\msq) model
to improve translation diversity within neural machine translation via
incorporating a set of discrete latent variables. We propose a model
architecture that requires negligible additional parameters and no
extra computation at inference time. In order to address prohibitive
memory requirement associated with large mixture models, we augment
the training procedure by computing the posterior distribution
followed by Monte Carlo sampling to estimate the gradients. We observe
significant gains both in terms of BLEU scores and translation
diversity with a mixture of $4$ components. In the future, we intend
to replace the uniform mixing coefficients with learnable parameters,
since different components should not necessarily make an equal
contribution to a given sentence pair. Moreover, we will consider
applying our \msq model to other NLP problems in which diversity
plays an important role.

\section{Acknowledgements}
We would like to thank Quan Tran, Trang Vu and three anonymous reviewers for their valuable comments and suggestions. This work was supported by the Multi-modal Australian ScienceS Imaging and Visualisation Environment (MASSIVE)\footnote{https://www.massive.org.au/}, and in part by an Australian Research Council grant (DP160102686).

\bibliography{bib}
\bibliographystyle{acl_natbib_nourl}

\onecolumn
\begin{appendices}
\section{ $n$-gram Diversity Measures}
\figref{fig:div} shows the unigram and bigram diversity comparison between the \msq\ model and the vanilla \sq\ model where the number of decoding beams is 8. 
Our \msq\ models significantly dominate the \sq\ model across language pairs in terms of these diversity metrics as well.

\begin{figure}[h]
    \centering
    \begin{subfigure}[b]{0.40\textwidth}
        \includegraphics[width=\textwidth]{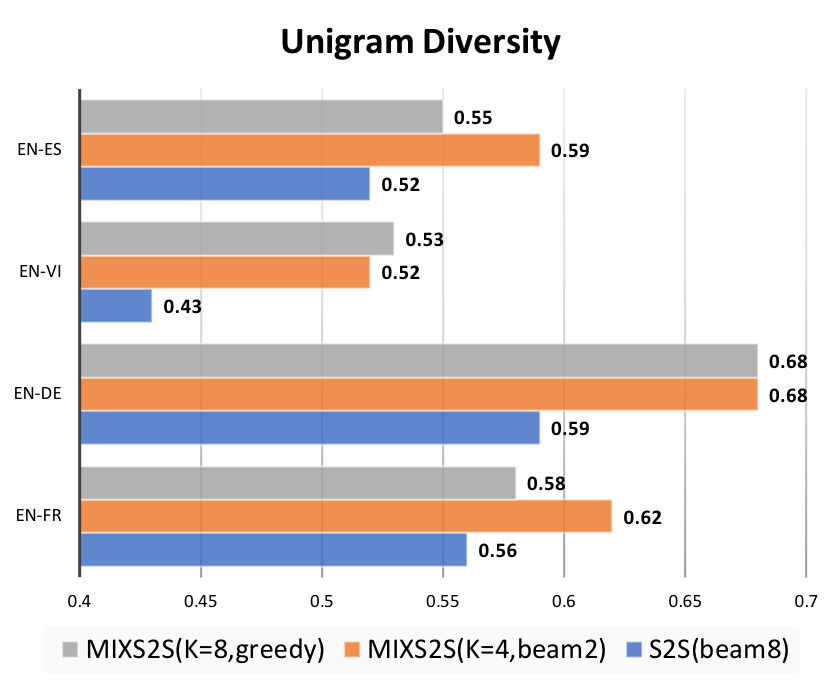}
    \end{subfigure}
    ~ 
    \begin{subfigure}[b]{0.40\textwidth}
        \includegraphics[width=\textwidth]{div_bi.png}
    \end{subfigure}
    ~ 
    \caption{unigram and bigram diversity plots of the main results}
    \label{fig:div}
\end{figure}

\section{Distribution Among Mixture Copmonents}
\begin{figure*}[h]
    \centering
    \begin{subfigure}[b]{0.45\textwidth}
        \includegraphics[width=\textwidth]{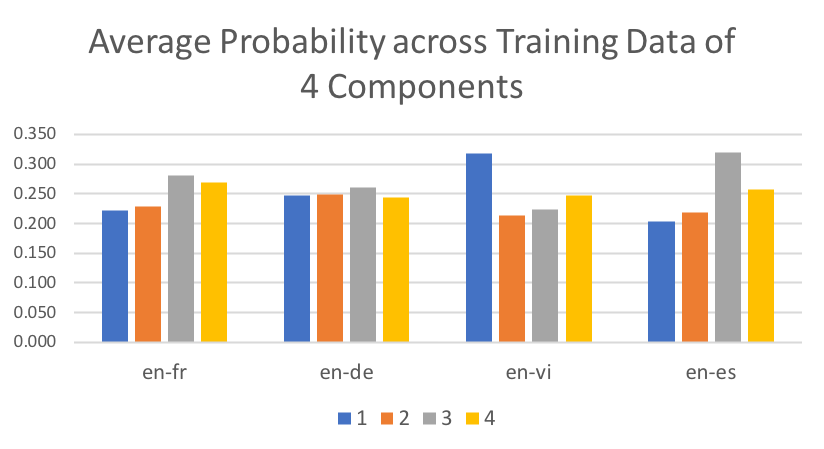}
    \end{subfigure}
    ~ 
    \begin{subfigure}[b]{0.45\textwidth}
        \includegraphics[width=\textwidth]{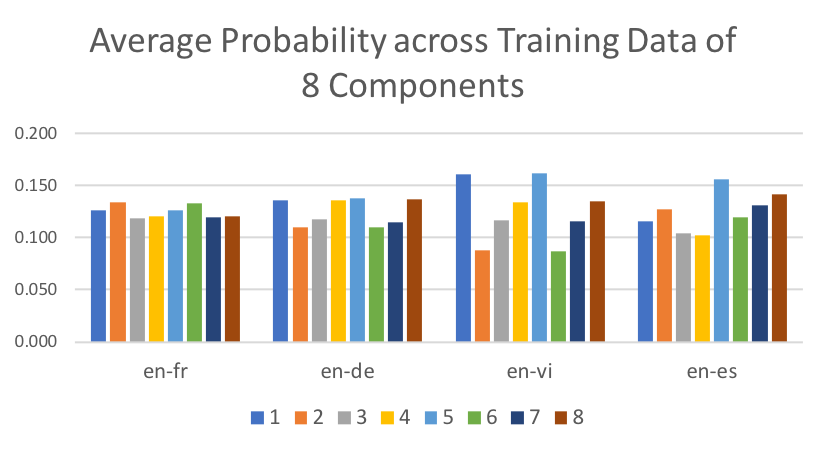}
    \end{subfigure}
    \begin{subfigure}[b]{0.70\textwidth}
        \includegraphics[width=\textwidth]{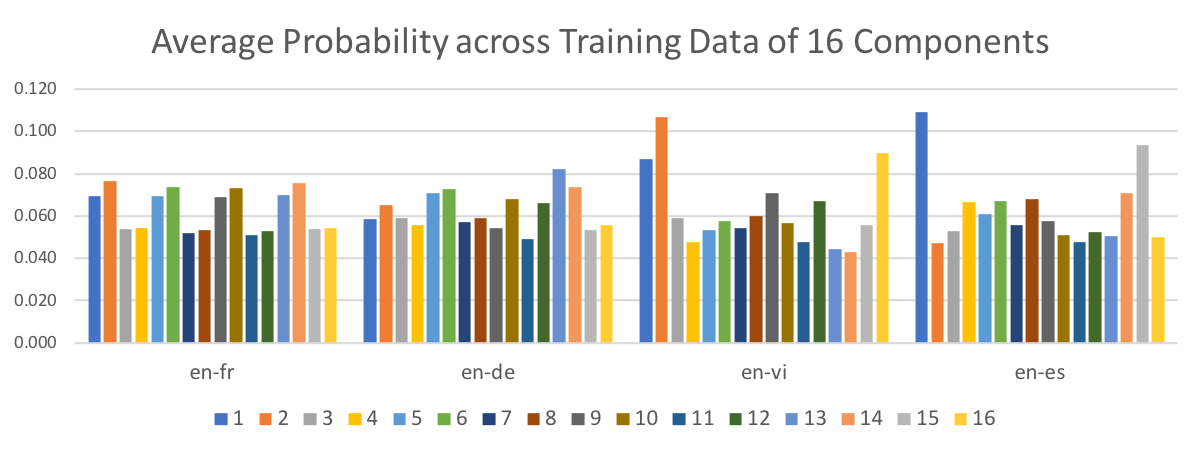}
    \end{subfigure}
    ~ 
    \caption{Average probability distribution across training data for different $K$}
    \label{fig:dist}
\end{figure*}

Due to the limitation of memory, we adopt Monte Carlo sampling to approximate $P(z\mid \bx,\by^*)$. We plot the average probability distribution across training corpora to visualize this approximation. According to \figref{fig:dist}, our approximation is not biased towards any mixture component, which is able to encourage diversity as we emphasized in the main paper. 

\section{Qualitative Examples}
\tabref{tab:supp-qual} show examples where \msq helped improve translation and demonstrate diversity, when compared with \sq.
\begin{table*}[hbt!]
\begin{center}
\begin{tabular}{ll||l}
\hline
\hline
Source & Talk about what you heard here . & I lost all hope .\\
Reference &Hãy \underline{kể} về những gì bạn \underline{được nghe} ở đây . &Tôi \underline{hoàn toàn tuyệt vọng} .\\
\sq&&\\
&Nói về những gì bạn vừa nghe . &Tôi \underline{mất hết hy} vọng .\\
&Nói về những gì bạn nghe thấy ở đây&Tôi \underline{mất tất cả hy} vọng\\
&Nói về những gì bạn nghe được ở đây&Tôi \underline{mất hết hy} vọng\\
&Nói về những gì bạn vừa nghe ở đây&Tôi \underline{đã mất hết hy} vọng\\

\msq&&\\
& \red{Hãy} \underline{nghe} về những gì bạn \underline{nghe được} \red{ở đây } . & Tôi \underline{mất tất cả hy} vọng .\\
& \red{Hãy} \underline{ban} về những gì bạn \underline{đã nghe} ở đây . & Tôi \underline{mất tất cả hy} vọng . \\
& \red{Hãy} \underline{bàn} về những gì bạn \underline{đã nghe} ở đây . & Tôi \underline{đã mất tất cả hy} vọng . \\
& \red{Hãy} \underline{bàn} về những gì bạn \underline{nghe được} ở đây .& Tôi \underline{mất tất cả hy} vọng .\\
\hline\hline
%
\end{tabular}
\end{center}
\caption{\underline{Words} indicate diversity compared with the references, while \red{red} words denote translation improvement.}
\label{tab:supp-qual}
\end{table*}

\end{appendices}

\end{document}